# 一种面向室内动态场景的 RGB-D SLAM 算法

苏登，丛德宏

(东北大学信息科学与工程学院，辽宁 沈阳 110819)

**摘 要**：视觉 SLAM 技术是机器人能够自主探索未知环境的关键技术之一，依据视觉传感器精确估计相机位姿是实现自主导航定位的基础，而绝大多数视觉 SLAM 算法基于静态环境假设，不能在动态环境中估计准确的相机位姿。为了解决这一问题，提出了一种应用于室内动态环境下的视觉 SLAM 算法。首先基于 RGB-D 相机的深度信息预先剔除部分运动物体，优化光度和深度误差获得初始相机位姿之后进一步剔除运动物体，之后将初步得到的静态背景重新用于位姿估计，多次迭代，在获得更为准确的静态背景的同时获得更为精确的相机位姿。试验结果表明，与前人研究成果相比，本文算法在低动态室内场景和高动态室内场景下均能获得更高的位姿估计精度。

**关键词**：视觉 SLAM（同时定位与地图构建）；动态环境；RGB-D 相机；优化；位姿估计

# A RGB-D SLAM Algorithm for Indoor Dynamic Scene

SU Deng, CHONG Dehong

(College of Information Science and Engineering, Northeastern University, Shenyang 110819, China)

**Abstract:** Visual slam technology is one of the key technologies for robot to explore unknown environment independently. Accurate estimation of camera pose based on visual sensor is the basis of autonomous navigation and positioning. However, most visual slam algorithms are based on static environment assumption and cannot estimate accurate camera pose in dynamic environment. In order to solve this problem, a visual SLAM algorithm for indoor dynamic environment is proposed. Firstly, some moving objects are eliminated based on the depth information of RGB-D camera, and the initial camera pose is obtained by optimizing the luminosity and depth errors, then the moving objects are further eliminated. and, the initial static background is used for pose estimation again. After several iterations, the more accurate static background and more accurate camera pose is obtained. Experimental results show that, compared with previous research results, the proposed algorithm can achieve higher pose estimation accuracy in both low dynamic indoor scenes and high dynamic indoor scenes.

## 1 引言（Introduction）

视觉 SLAM 技术对移动机器人自主导航定位有着重要作用。利用视觉传感器恢复移动机器人的位姿，重建三维环境是视觉 SLAM 的核心问题，RGB-D 相机相比单目或双目相机，能够低成本地获取像素的深度，并提供环境的彩色信息，在视觉 SLAM 中得到了广泛应用。

现阶段绝大多数视觉 SLAM 都基于静态场景的假设，视觉传感器采集的变化信息仅依赖于相机的运动，文献[1-3]是经典的静态场景视觉 SLAM 算法。然而实际环境下很难满足绝对静态这一假设，场景中不可避免的会出现运动物体例如走动的行人。基于静态假设的视觉 SLAM 算法在面对动态场景时其精度和鲁棒性将显著降低。ORB-SLAM2[4]利用数据的关联特性，能够较好的适应低动态场景，但在高动态场景下，其相机位姿估计会出现较大漂移。

面向动态场景的视觉 SLAM 问题相比静态场景主要增加运动物体的检测和剔除环节，目前视觉 SLAM 中运动分割的方法可主要分为以下四种：基于先验知识的方法，基于约束的方法，基于光流的方法，基于深度学习的方法。

文献[5]基于对运动物体的先验信息剔除运动物体上的特征点。与之类似地，文献[6]将运动物体上的 SURF 特征描述子储存之后，通过与每一帧中提取的特征点描述子对比来检测剔除运动物体。

基于几何约束的方法利用对极几何分离静态特征和动态特征。其基本原理是动态特征不满足基于静态特征的多视图几何约束。Kundu 等人[7]通过构造本质矩阵定义两种几何约束：对极几何要求匹配点对应的空间点处于两条极线的相交点处，若相距过远，则视为动态特征点；对于特征点沿极线运动的情况，采用 FVB 约束，在被跟踪特征的流上设置上下边界，位于边界外的被跟踪特征将被检测为移动特征。文[8]利用三角化原理区分静态特征和动态特征，但容易在低纹理和光照变化剧烈的区域丢失特征点，也容易受到传感器噪声的影响。

光流法通过连帧图像间的亮度定义当前运动[9]。通常用来匹配图像中的运动区域来分割运动物体。Klappstein[10]定义由光流计算的运动矩阵来衡量物体运动的可能性，如果场景中存在运动物体，运动矩阵将评估其在多大程度上违反了光流。Alcantarilla 等人[11]根据 3D 运动矢量在场景流（光学流的 3D 版本）中的模数，通过剩余运动似然来分割运动对象。

传统 SLAM 问题通过八点法或者五点法计算相机运动，这类传统方法未对相机的运动模式有所限制，另一种方法就是对相机运动加上限制信息，例如车轮里程信息，这样将符合相机运动约束的特征视为静态特征。Scaramuzza[12]提出使用轮式车辆的非完整约束来计算相机运动，基于相机的运动只能是二维平面运动的假设

建立模型，利用这一假设，相机的运动便只有一个自由度，可以使用一对点求解[13]。

近年来，深度神经网络 DNN(Deep Neural Networks) 在计算机视觉中获得了广泛的欢迎。尽管已经出现了用于视觉定位和 3D 重建的多种实现方式，但用于运动分割的 DNN 方法仍然很少。目前来说，DNN 主要用来配合以上的方法进行运动分割。

部分学者融合多种算法用于处理复杂多变的环境，文[14]采用直接法估计相机的初始位姿，然后通过特征点匹配和最小化重投影误差进一步优化位姿，通过筛选地图点并优化位姿输出策略，使算法能够处理稀疏纹理、光照变化、移动物体等环境；文[15]通过最小化图像光度误差，利用稀疏图像对齐算法实现对相机位姿的初步估计,依据高斯模型提出运动特征后，通过最小化重投影误差对相机位姿进行进一步优化，提升相机位姿估计精度.

本文算法提出了一种融合动态剔除的视觉里程计，在实际操作过程中，利用深度信息对动态物体进行预剔除后，联合优化深度误差与光度误差进一步剔除动态物体，将更新后的静态背景重新输入到视觉里程计中优化求解相机位姿，多次迭代后获得收敛的更为精确的精确相机位姿。本文算法在 TUM RGB 数据集[16]和实际环境中进行了测试，并在多个指标上与已有先进算法进行了比较，实验结果证明本文算法在面向室内动态环境时具有更好的精度和鲁棒性。

## 2 系统流程框架（Frame and process of the system）

本文算法流程框架如下图 1 所示，输入两对 RGB-D 相机采集的图像，每对图像包含一张光度图以及一张深度图，A，B 的深度图与光度图一一对应。通过深度差异预先剔除部分运动物体之后，最小化深度和光度误差，得到初始相机位姿，将 B 中的光度图通过位姿变换后与 A 中的光度图计算残差获得离群值，将离群值视为运动物体，剔除运动物体获得静态背景图像，再将背景图像返回至优化模块进行多次迭代，获得更为准确的相机位姿。

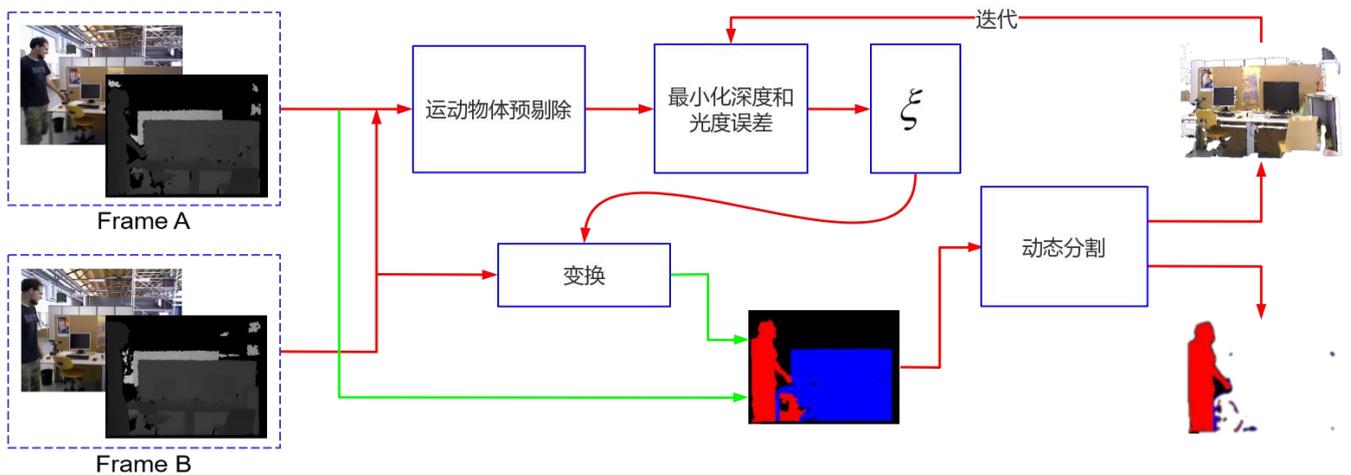

图 1 系统流程图

Fig.1 Flow chart of the system

## 3 运动物体检测与预剔除（Moving object detection and pre elimination）

RGB-D 相机能够低成本地获取光度图像及其对应像素的深度，本文算法以 RGB-D 相机作为唯一外部传感器，仅以深度信息预先剔除动态物体。

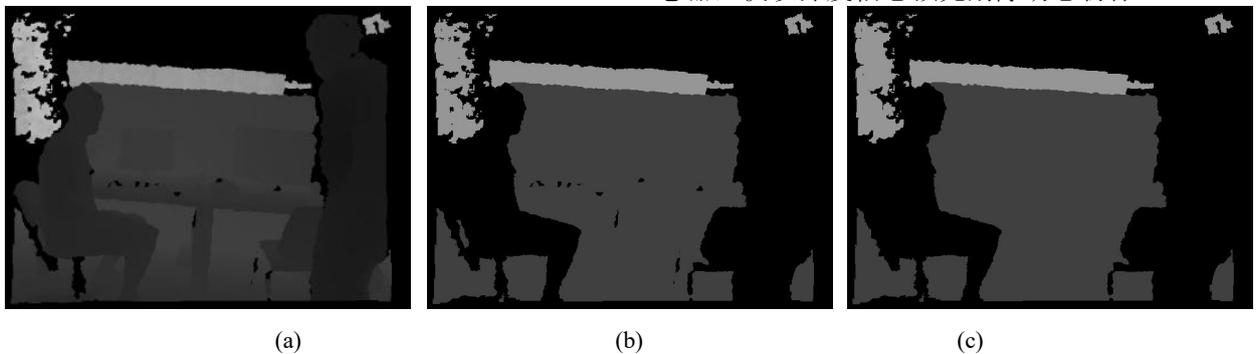

(a)　　　　　　　　　(b)　　　　　　　　　(c)

图 2 深度聚类

Fig.2 Depth clustering

首先，通过 $k$ 均值聚类将深度图像中的物体通过深度信息聚类为 $n$ 类，如图 2 所示，图(a)为未聚类深度图像，图(b)是将图(a)聚类数取值为 5 后的深度图像。由于深度相机自身的数据噪声，导致原始深度图像和聚类后深度图像中存在空洞，在此采用形态学重构原理填补这些空洞，结果如图(c)所示。

可以看到在聚类后，静态背景与可能产生运动的行人在深度上有明显划分，在这下一步通过对比深度变化

最大限度剔除运动物体极为重要。

在没有先验知识的条件下，聚类数的选取采取如下方法：

$$n_{cluster} = \left\{ \frac{D_{max} - D_{min}}{D_{sd}} \right\} \quad (1)$$

其中 $n_{cluster}$ 代表聚类数，$D_{max}$、$D_{min}$、$D_{sd}$ 分别代表最大深度，最小深度和深度标准差，符号{.}在这里定义为不超过符号内数值的最大正整数。

将每张原深度图像划分为大小相同的 $k$ 个区域，若第 $k_i$ 个区域内存在动态物体，则两帧深度图对应的该区域内必将存在深度变化。为了区分深度变化是由相机自身运动造成的还是运动物体造成的，在对比深度变化时，忽略只存在某一深度的区域，关注包含多个深度的区域，换言之就是关注包含物体"边缘"的区域。如图3所示，对于所有包含"边缘"的区域而言，如果深度变化仅由相机运动产生，那么区域中"边缘"两侧的深度差值必将符合如下约束：

(1)若相机左右平移运动或前后平移运动，则两帧深度图像所有区域"边缘"两侧深度差值不变，如(a)所示，当相机发生前后平移或左右平移时，在深度图像中红色静止物体与背景的距离 ab，a1b1 不会发生改变；

(2)若相机发生旋转，如(b)所示：在初始位置下，两个静止物体距背景的距离分别为 ac，a1c1；当相机和初始位置相比发生了一个平移和旋转后，三角形 abc 与三角形 a1b1c1 相似，故 ab 与 ac 的比值等于 a1b1 与 a1c1 的比值，在此种情况下，相机运动前后静止物体与背景的距离符合正比例关系。

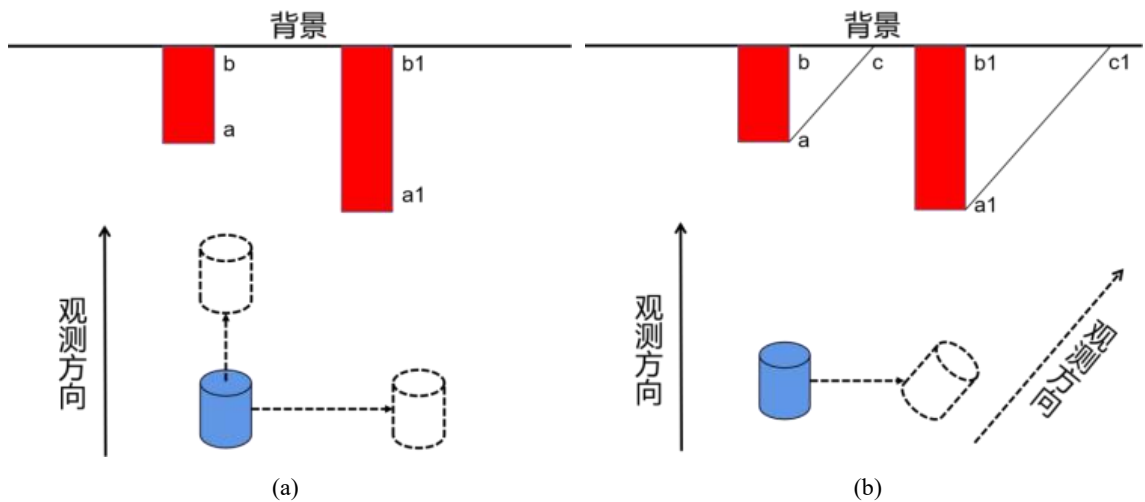

图 4 深度约束关系

Fig.4 Relationship of depth constrain

在此基础上视不符合特定差值变化的区域为包含动态物体的区域，且通过此方法可以判断深度差值是由哪一部分物体造成的。先前通过聚类操作使得图像中的深度分别聚类，此时将这一物体所属的最大连通区域聚类深度认为是运动物体所在深度并剔除，例如行人在相机前走过时，某连续两帧深度图像之间能够通过深度变化判断为运动物体的仅仅是行人四肢和躯干的边缘部分，而剔除聚类后的最大连通深度区域能在最大程度上剔除掉整个行人。但是在非理想情况下，行人不一定会聚类成同一深度，因此在进行运动物体预剔除之后，本文采取优化的方法多次迭代更新静态背景，进一步剔除运动物体，获得更精确的位姿。

## 4 位姿估计（Pose estimation）

因为观测噪声的存在，观测方程中存在一个噪声项，假设噪声符合零均值高斯分布，则观测方程可以表示为以下的形式：

$$z_{k,j} = f(l_j, x_k) + w_{k,j} \quad (2)$$

其中 $z_{k,j}$ 为观测数据，具体表现为像素坐标系下的深度和光度数据，$f(\bullet)$ 为观测方程，$w_{k,j}$ 为零均值分布的高斯噪声，且 $w_{k,j} \sim N(0, Q_{k,j})$，则观测数据的条件概率为：

$$P(z_{k,j} | l_j, x_k) = N(f(l_j, x_k), \mathbf{Q_{k,j}}) \quad (3)$$

为求其最大值，对其取负对数，等价于求负对数最小值：

$$argmin((z_{k,j} - f(l_j, x_k))^T \mathbf{Q_{k,j}}^{-1} (z_{k,j} - f(l_j, x_k))) \quad (4)$$

在光度图中，$z_{k,j} - f(l_j, x_k)$ 为光度误差，在深度图中为深度误差。进一步地，将协方差矩阵 $\mathbf{Q_{k,j}}$ 视为单位矩阵，在进行运动物体预剔除后，视相机位姿为优化变量，光度误差和深度误差为联合优化目标函数，构建非线性优化问题：

$$\xi = argmin_\xi \left\{ \sum_{P=1}^{N} \left\| \alpha_I r_I^P(\xi) + r_D^P(\xi) \right\|^2 \right\} \quad (5)$$

其中 $\xi$ 为相机位姿，$r_I^P$ 和 $r_D^P$ 分别代表 $P$ 区域内光度误差和深度误差。由于光度和深度单位不同，所以引入尺度因子 $\alpha_I$。

光度误差或深度误差的计算不是简单的在两帧图像中选取相同像素坐标的像素点相减，因为随着相机运动，对应点在两帧图像中的像素坐标必然发生变化，对比光度和深度误差方法如图 5 所示：

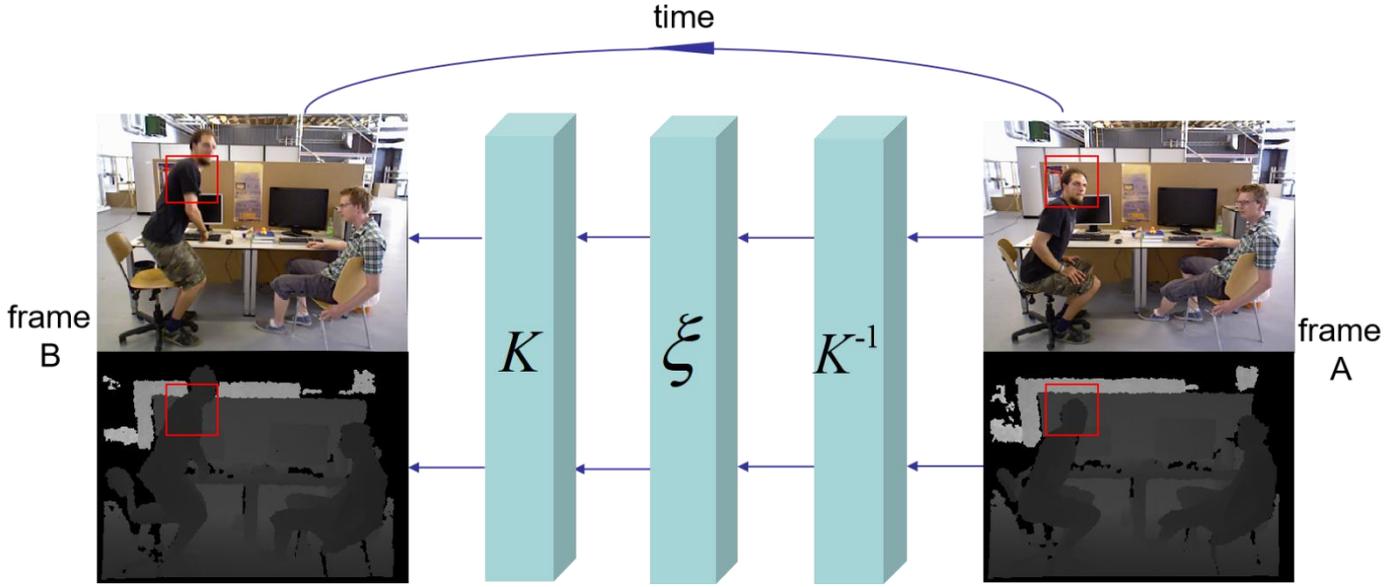

图 5 像素坐标转换示意图

Fig.5 Schematic diagram of pixel coordinate conversion

对比标记红色方框中的光度和深度误差应先通过相机内参数将像素坐标恢复到相机坐标系下的空间坐标，再通过相机位姿变化转换到下一状态相机坐标系坐标，最后投影到像素坐标系下进行比较：

$$r_I^P(\xi) = I_B\left(\omega(x^P,\xi)\right) - I_A\left(x^p\right) \quad (6)$$

$$r_D^P(\xi) = D_B\left(\omega(x^P,\xi)\right) - \left|T(\xi)K^{-1}(x^P, D_A(x^P))\right|_D \quad (7)$$

其中 $x^P$ 为 A 帧 P 区域中某个像素点，$I(\cdot)$ 为取某点光度，$D(\cdot)$ 为取某点深度，K 为相机内参数，一般已经提前标定准确。$\omega(\cdot)$ 为像素点在两种相机位姿下像素坐标转换函数，形式如下：

$$\omega(x^P,\xi) = KT(\xi)K^{-1}(x^P, D_A(x^P)) \quad (8)$$

$T(\xi)$ 为李代数形式的相机位姿 $\xi$ 通过指数映射在李群下的表现形式。对于构建的非线性优化问题，本文算法采用高斯-牛顿法(Gauss-Newton)构建正规方程(Normal equation)迭代求解，在上一节中预先剔除动态物体的边缘部分可以有效防止由运动物体边缘造成的局部过大光度和深度误差，以免在方程迭代求解过程中陷入局部最优解。

## 5 背景更新（Background update）

由于基于深度聚类的运动物体预剔除不能完全保证剔除了所有运动部分，处于非边缘深度上且未聚类为同一类的运动部分将得到保留，之后通过优化的方法得到的相机位姿和静态背景距离真实值依旧存在一定误差。本文算法在获得初始静态背景后，将其返回至优化模块，由于第一次优化进一步提出了部分运动物体，第二次优化时将获得更精确的相机位姿，并再一次剔除部分运动物体。本文设置迭代次数七次，图 6 是连续六次迭代后运动物体的剔除情况，可以看到在经历四次迭代后，运动物体已基本剔除干净，第五次和第六次迭代并未有太大变化，说明迭代数值已经收敛。

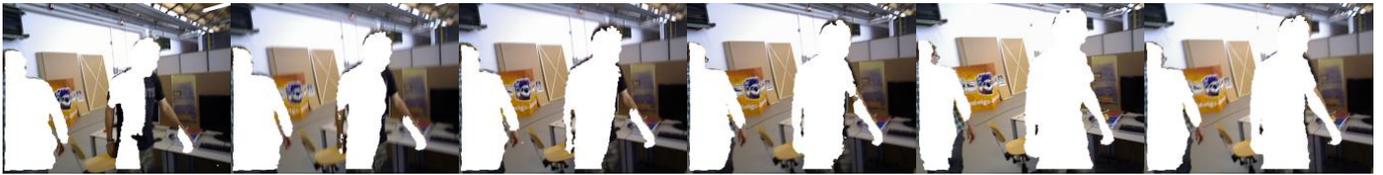

图 6 迭代优化剔除运动物体

Fig.6 Eliminating moving objects by iterative optimization

## 6 实验（Experiment）

本文算法在实际环境和 TUM RGB-D 数据集下对本文算法进行了测试。TUM 数据集中基于动态场景的 RGB-D 序列广泛应用于动态场景下视觉 SLAM 系统的性能测试。TUM 中将动态场景数据集分为"walking"和"sitting"系列，在静态背景一致的情况下，"walking"系列中的动态部分主要是移动的行人，包含"起立"，"坐下"，"行走"，"原地旋转"等复杂运动，且占据较大的视角；"sitting"系列中的动态部分仅为人物的肢体和一些小物体，动态部分占据视角较小。实验部分本文算法与 ORB-SLAM2，DVO(dense visual odometry and SLAM)[17]，BaMVO(effective background model-based RGB-D dense visual SLAM)[18]等经典算法和文[19]中的方法进行了对比。

程序运行环境为 Intel Core i7-8750H CPU，2.20GHz 主频，7.7GiB 内存，不使用 GPU 加速。运行系统为 Ubuntu18.04.5 LTS。

### 6.1 动态物体剔除

本文算法在深度聚类后将深度图像划分为 N 个区域，加权对比深度变化后评估某区域的动态系数，将动

态系数大于 0.5 的区域以及与其归为同一类深度的连通区域视为动态物体并剔除，这一方法能够排除光照变化的影响。对于深度图像在物体边缘出现的空洞现象，采用向右临近深度补齐的原理消除边缘零值深度噪声。

图 7 为本文算法在 TUM 数据集以及实际环境中几张具有代表性的动态物体剔除结果。

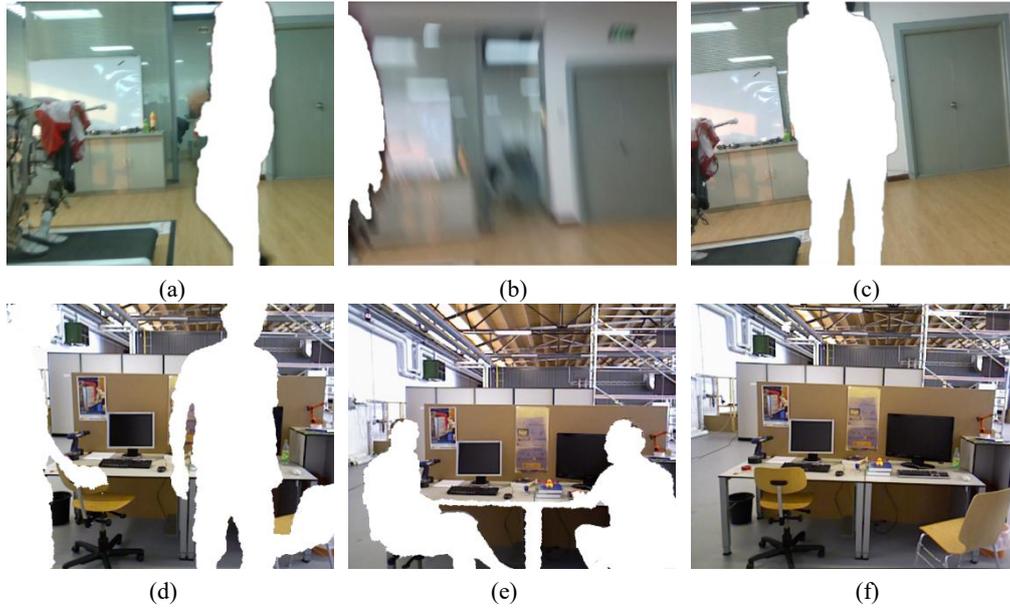

图 7 运动物体剔除效果

Fig.7 The effect of moving object elimination

图 7 展示了不同情况下的运动物体剔除情况：图(a)和(d)中实际场景和数据集镜头前有快速运动的行人，且占据较大的视角，可以看到行人的轮廓几乎完全剔除，在实际场景中虽然左侧机器人和行人聚类为同一深度，但两者不存在连通区域，因此在剔除行人时保留了左侧机器人，在数据集中由于行人与椅子的一部分在深度上聚类为同一类，且两者存在连通区域，因此在剔除行人的同时剔除了椅子的一部分，但随着行人的运动以及相机的运动，两者不可能永远处于同一深度聚类上，在三维重建中依旧可以恢复出完整的椅子；图(b)中相机镜头存在轻微晃动，与上一帧相比由相机运动造成的背景深度变化显著，仍然在很大程度上剔除了左侧移动的行人；图(c)与上一帧比较，行人只有手部的轻微晃动，图(e)中两人只有头部和腿部的轻微运动，但依旧将他们的轮廓完全剔除。图(f)中没有运动的物体，与上一帧比较只有相机运动造成的深度变化，因此没有物体被剔除。

可以看到本文算法在高动态场景下和低动态场景下均能够获得良好的运动物体剔除效果，且能够排除相机运动造成的深度变化干扰，尤其是在相机发生晃动时，也能够在极大程度上剔除运动物体，在后续进行位姿估计时大大减小运动物体对位姿估计的影响。以上结果证明本文算法对运动物体的剔除具有良好的鲁棒性。

**6.2 视觉里程计评估**

本文算法根据深度变化预先剔除运动物体后，优化光度和深度误差于连续两帧 RGB 图像之间恢复出帧间相机位姿变换。本节与 TUM 数据集提供的真实轨迹对比，采用相对位姿误差(relative pose error,PRE)评估本文算法设计的视觉里程计在动态环境下的位姿输出精度，并于 DVO 等算法进行比较。结果如表 1 所示。

表 1 相对位姿误差对比

Tab.1 Comparison of relative pose errors

| | 图像序列 | 平移 RMSE/(m/s) | | | | 旋转 RMSE/(°/s) | | | |
|---|---|---|---|---|---|---|---|---|---|
| | | DVO | BaMVO | 文献[19] | 本文 | DVO | BaMVO | 文献[19] | 本文 |
| 低动态 | fr2/desk_peson | 0.0354 | 0.0352 | **0.0069** | 0.0099 | 1.5368 | 1.2159 | 0.4380 | **0.4247** |
| | fr3/sitting_static | 0.0157 | 0.0248 | **0.0077** | 0.0083 | 0.6084 | 0.6977 | **0.2595** | 0.2677 |
| | fr3/sitting_xyz | 0.0453 | 0.0482 | **0.0117** | 0.0173 | 1.4980 | 1.3885 | 0.4997 | **0.4980** |
| | fr3/sitting_rpy | 0.1735 | 0.1872 | 0.0234 | **0.0233** | 6.0164 | 5.9834 | 0.7838 | **0.7134** |
| | fr3/sitting_halfsphere | 0.1005 | 0.0598 | **0.0245** | 0.0463 | 4.6490 | 2.8804 | **0.5643** | 0.9621 |
| 高动态 | fr3/walking_static | 0.3818 | 0.1339 | 0.1881 | **0.0306** | 6.5302 | 2.0833 | 3.2101 | **0.5882** |
| | fr3/walking_xyz | 0.4360 | 0.2326 | 0.2158 | **0.1006** | 7.6669 | 4.3911 | 3.6476 | **1.9498** |
| | fr3/walking_rpy | 0.4038 | 0.3584 | 0.3270 | **0.1415** | 7.0662 | 6.3398 | 6.3215 | **2.3033** |
| | Fr3/walking_halfsphere | 0.2638 | 0.1738 | 0.1908 | **0.1121** | 5.2179 | 4.2863 | 3.3321 | **2.1003** |

可以看到，无论在在低动态场景条件下，还是高动态场景条件下，本文算法的平移方均根误差和旋转方均根误差相较于 DVO 与 BaMVO 有极大幅度的降低。

在低动态场景条件下，相比于 DVO，本文算法平移方均根误差最多降低了 86.6%，最少降低了 47.1%；旋转方均根误差最多降低了 88.1%，最少降低了 56.0%；相比于 BaMVO，本文算法平移方均根误差最多降低了 87.6%，最少降低了 22.6%；旋转方均根误差最多降低了 88.1%，最少降低了 61.1%。在低动态场景条件下，和文献[19]相比本文算法在平移精度和旋转精度上各有侧重，但误差数值非常接近。

在高动态场景下，本文算法在平移精度上和旋转精度上相较于 DVO，BaMVO 和文献[19]都有极大程度的提高：在四种高动态数据集下，平移方均根误差相较于 DVO，BaMVO 和文献[19]最多分别降低了 92.0%，77.1%，83.7%，最少分别降低了 57.5%，35.5%，41.2%；旋转方均根误差最多分别降低了 91.0%，71.6%，81.7%，最少分别降低了 59.7%，51.0%，37.0%。

DVO 算法基于静态背景的假设使得其在仅存在微小运动的"sitting"序列和 fr2/desk_peson 数据集中的平移与旋转方均根误差较小，但在"walking"系列下，由于大幅度运动物体的存在，使其平移与旋转方均根误差显著增大。与 DVO 算法相比，BaMVO 算法根据运动物体边缘深度变化，剔除了一部分运动特征点，但也必然将保留运动物体非边缘处的运动特征，对相机位姿估计造成干扰。由于对部分运动特征点进行了剔除，BaMVO 算法在低动态和高动态场景下精度相较于 DVO 算法有所提升，但依旧存在很大的误差，两者均不能适应高动态环境。文献[19]中算法采用几何约束剔除动态特征点，但由于特征点的提取在每一次实验中都不是固定点，因此文献[19]在预先剔除部分特征点后，仍然存在在恢复相机位姿的过程中检测出新的运动特征点的情况，这在高动态场景下表现得较为明显：相较于 BaMVO 算法，其平移方均根误差在数据集 fr3/walking_static 和 Fr3/walking_halfsphere 上分别高出 40.5%和 9.8%，旋转方均根误差在数据集 fr3/walking_static 上高出 53.7%。本文算法和文献[19]相比在低动态场景条件下的平移精度和旋转精度上各有侧重，但误差数值非常接近，唯一例外出现在 fr3/sitting_halfsphere 数据集中，因为本文算法将微小运动物体及其同一深度的连通区域一并剔除，使得在某些帧中出现了特征丢失的情况：在这些帧中微小运动物体所属的完整个体本身包含了本帧绝大多数特征；在高动态场景条件下，本文算法相较于表 1 中其它三种算法在四个数据集下都取得了最小平移方均根误差和旋转方均根误差，且精度提高显著。

**6.3 SLAM 系统评估**

SLAM 系统在前端视觉里程计估计相机位姿之后，在后端通过优化和回环检测的方法进一步消除前端位姿估计造成的累积误差。本节使用绝对轨迹误差(absolute trajectory error，ATE)评估 SLAM 系统的精度，绝对轨迹误差直接计算相机位姿的真实值与 SLAM 系统的估计值之间的差，根据位姿的时间戳将真实值和估计值进行对齐，然后计算每对位姿之间的差值，非常适合 SLAM 系统的评估。本节将通过与文献[19]和 ORB-SLAM2 的绝对轨迹误差进行对比，评估本文算法的精度。

本文算法在三个具有代表性的数据集 (fr3/sitting_xyz, fr3/walking_xyz, Fr3/walking_halfsphere) 下与 ORB-SLAM2 进行对比，如图 8 所示。

图 8 中(a)，(b)，(c)分别代表 ORB-SLAM2 在 fr3/sitting_xyz, fr3/walking_xyz 和 Fr3/walking_halfsphere 下的绝对轨迹误差情况；(d)，(e)，(f)代表本文算法在对应数据集下的绝对轨迹误差情况。ORB-SLAM2 在前端采用提取 ORB 特征点的方法恢复相机位姿，在后端通过重投影误差剔除部分离群值，在低动态场景下能够获得较高精度，但同样不能适应高动态场景条件。由图 8 可以看出，本文算法在低动态场景条件下和 ORB-SLAM2 均具有非常高的精度，而在高动态场景条件下，本文算法精度相较于 ORB-SLAM2 有大幅度提升。

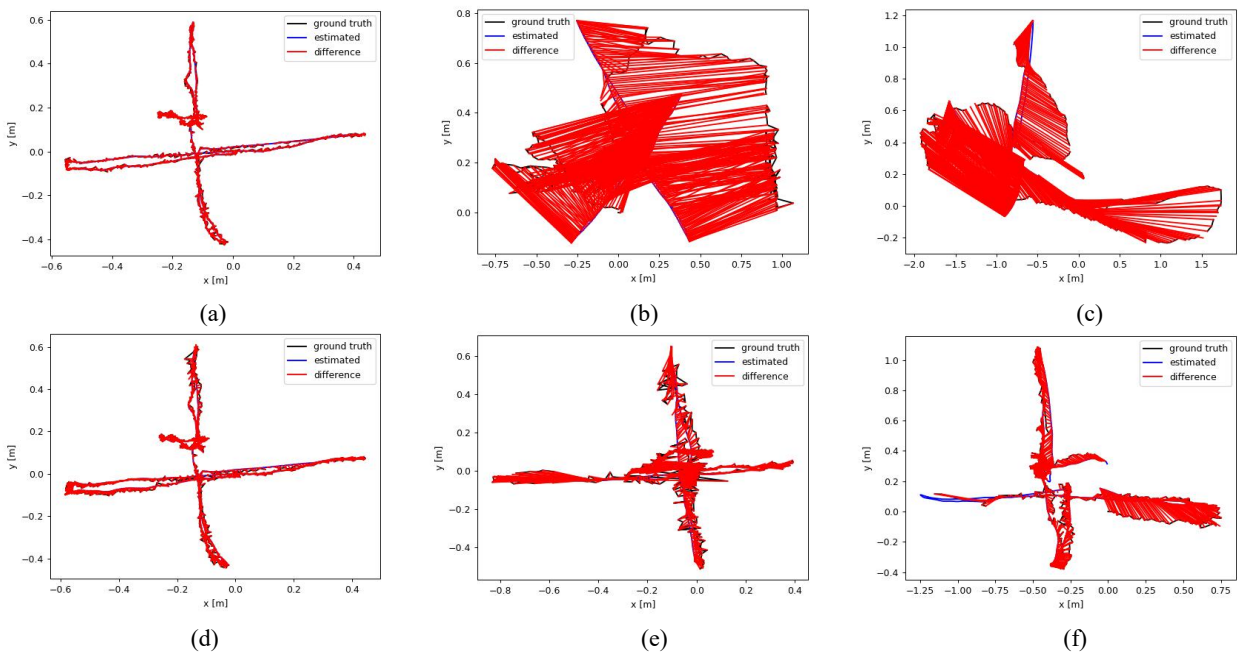

图 8 绝对轨迹误差对比图

Fig.8 Comparison chart of absolute trajectory errors

表 2 绝对轨迹误差对比表

Tab.2 Comparison table of absolute trajectory error

| 图像序列 | ORB-SLAM2 ATE(m) | | | 文献[5] ATE(m) | | | 本文算法 ATE(m) | | |
|---|---|---|---|---|---|---|---|---|---|
| | Max | Min | RMSE | Max | Min | RMSE | Max | Min | RMSE |
| fr3/walking_static | 0.6506 | 0.0646 | 0.3902 | 0.6161 | 0.0231 | 0.3080 | **0.1523** | **0.0007** | **0.0218** |
| fr3/walking_xyz | 1.4540 | 0.0590 | 0.6731 | 0.8190 | 0.0430 | 0.3047 | **0.4584** | **0.0034** | **0.0799** |
| fr3/walking_rpy | 1.7196 | 0.0285 | 0.8004 | 0.9351 | 0.0708 | 0.4983 | **0.4678** | **0.0171** | **0.1510** |
| Fr3/walking_halfsphere | 1.6433 | 0.0463 | 0.5571 | 0.6246 | 0.1088 | 0.3116 | **0.4065** | **0.0057** | **0.1147** |

表 2 通过方均根绝对轨迹误差，最大绝对轨迹误差和最小绝对轨迹误差三个指标定量在高动态场景下比较了文献[19]，ORB-SLAM2 和本文算法。

对比 ATE 可知，本文算法在四个高动态数据集下的三个指标上均优于 ORB-SLAM2 和文献[19]，以数据集 fr3/walking_xyz，本文算法的最大绝对轨迹误差，最小绝对轨迹误差和方均根绝对轨迹误差相较与 ORB-SLAM2 分别减少了 68.5%，94.2%，88.1%；相较于文献[19]分别减少了 44.0%，92.1%，73.8%。

### 6.4 地图重建

视觉 SLAM 问题本质上可以看作相机位姿估计问题，能否准确地重建三维地图，关键在于能否准确地获得相机位姿。本节对 TUM 高动态场景数据集 fr3/walking_xyz 构建点云地图，结果如图 9 所示。可以看到在剔除动态物体之后，静态背景均没有出现撕裂，扭曲，交叠的情况，重建结果中出现的空白区域是相机未捕捉到的区域。可见本文算法通过动态物体的剔除，与真实轨迹相比以较高精度获得了相机位姿，并在动态场景的重建过程中保留了绝大部分静态背景。

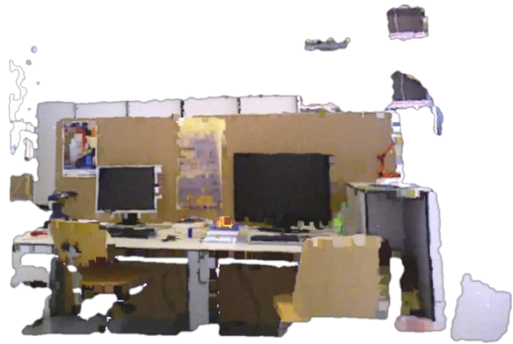

图 9 地图重建结果

Fig.9 Results of map reconstruction

## 7 结论（Conclusion）

本文提出了一种面向室内动态环境的 RGB-D SLAM 算法。本文算法利用 K 均值聚类深度信息，利用边缘深度变化检测运动物体，并寻找最大连通聚类深度预剔除运动物体，接着通过优化的方法获得初始相机位姿和静态背景，最后采取迭代的方式更新静态背景，获得精确的相机位姿。本文算法在 TUM 数据集和实际环境下进行了检测，并于其他面向动态场景的 SLAM 算法进行了比较，实验结果证明本文算法在低动态场景和高动态场景下均能获得更为精确的相机位姿。下一步工作将结合深度学习，在光照条件不稳定的室内动态场景中提升位姿估计精度和系统鲁棒性。